\documentclass[preprint,conference]{IEEEtran}
\IEEEoverridecommandlockouts
\usepackage{cite}
\usepackage{amsmath,amssymb,amsfonts}
\usepackage{algorithmic}
\usepackage{graphicx}
\usepackage{textcomp}
\usepackage{xcolor}
\PassOptionsToPackage{hyphens}{url}
\usepackage{url}
\def\BibTeX{{\rm B\kern-.05em{\sc i\kern-.025em b}\kern-.08em
T\kern-.1667em\lower.7ex\hbox{E}\kern-.125emX}}

\makeatletter
\newcommand{\linebreakand}{%
\end{@IEEEauthorhalign}
\hfill\mbox{}\par
\mbox{}\hfill\begin{@IEEEauthorhalign}
}
\makeatother

\usepackage{subfigure}

\usepackage[utf8]{inputenc}
\usepackage[english]{babel}
\usepackage{url}
\usepackage{graphicx}
\usepackage{subfig}
\usepackage{pgfplotstable}
\usepackage{pgfplots}
\usepackage{booktabs}
\usepackage{etoolbox,lipsum}
\usepackage{comment}
\usepackage{siunitx}
\usepackage{enumitem}

\usepackage{tikz}
\usetikzlibrary{shapes,arrows,shadows}
\usetikzlibrary{mindmap,trees}
\definecolor{mygreen}{rgb}{0.553,0.682,0.063}
\definecolor{myblue}{rgb}{0.0,0.208,0.376}
\definecolor{mygray}{rgb}{0.906,0.906,0.906}
\definecolor{uni_oldenburg}{HTML}{0d59ab}

\definecolor{darkblue}{rgb}{0.0,0.0,0.8}
\definecolor{mydarkgreen}{rgb}{0.0,0.5,0.0}
\definecolor{tablehighlight}{rgb}{0.0,0.5,0.0}

\newcommand*{\mycolorbox}[1]{%
\tikzstyle{mybox} = [draw=black, rectangle, inner sep=0.5pt, inner ysep=1pt, inner xsep=1pt, fill=white]
\tikzstyle{fancytitle} = [fill=white, text=black]
\begin{tikzpicture}
\node [mybox] (box){%
#1
};
\end{tikzpicture}%
}

\newcommand{\ie}{i.\nolinebreak[4]\hspace{0.01em}\nolinebreak[4]e.}
\newcommand{\eg}{e.\nolinebreak[4]\hspace{0.01em}\nolinebreak[4]g.}
\newcommand{\etal}{\emph{et al.}}

\usepackage{amsfonts}       
\usepackage{amsmath, bm}    

\newcommand{\x}{\bm{x}}
\newcommand{\y}{y}

\newcommand{\X}{X}
\newcommand{\Y}{Y}

\newcommand{\model}{f}

\newcommand{\npattern}{n}
\newcommand{\nclasses}{c}

\newcommand{\iwidth}{w}
\newcommand{\iheight}{h}

\newcommand{\ndim}{d}

\newcommand{\noroad}{\texttt{no road}}
\newcommand{\smallroad}{\texttt{small road}}
\newcommand{\mediumroad}{\texttt{medium road}}
\newcommand{\bigroad}{\texttt{big road}}
\newcommand{\tverskyloss}{\mathcal{L}}

\newcommand{\unet}{\texttt{U-Net}}
\newcommand{\unetplus}{\texttt{U-Net+}}
\newcommand{\unetplusshift}{\texttt{U-Net+(shift)}}
\newcommand{\unetplustimeflat}{\texttt{U-Net+Time(flat)}}
\newcommand{\unetplustimevolumetric}{\texttt{U-Net+Time(3d)}}

\begin{document}

\title{Detecting Hardly Visible Roads in\\Low-Resolution Satellite Time Series Data}

\author{
\IEEEauthorblockN{Stefan Oehmcke}
\IEEEauthorblockA{\textit{Dep. of Computer Science} \\
\textit{University of Copenhagen}\\
Copenhagen, Denmark\\
stefan.oehmcke@di.ku.dk}
\and
\IEEEauthorblockN{Christoffer Thrys{\o}e}
\IEEEauthorblockA{\textit{Dep. of Computer Science} \\
\textit{University of Copenhagen}\\
Copenhagen, Denmark\\
dfv107@ku.alumni.dk}
\and
\IEEEauthorblockN{Andreas Borgstad}
\IEEEauthorblockA{\textit{Dep. of Computer Science} \\
\textit{University of Copenhagen}\\
Copenhagen, Denmark\\
pmh477@ku.alumni.dk}
\linebreakand
\IEEEauthorblockN{Marcos Antonio Vaz Salles}
\IEEEauthorblockA{\textit{Dep. of Computer Science} \\
\textit{University of Copenhagen}\\
Copenhagen, Denmark\\
vmarcos@di.ku.dk}
\and
\IEEEauthorblockN{Martin Brandt}
\IEEEauthorblockA{\textit{Dep. of Geosciences} \\
\textit{University of Copenhagen}\\
Copenhagen, Denmark\\
mabr@ign.ku.dk}
\and
\IEEEauthorblockN{Fabian Gieseke}
\IEEEauthorblockA{\textit{Dep. of Computer Science} \\
\textit{University of Copenhagen}\\
Copenhagen, Denmark\\
fabian.gieseke@di.ku.dk}
}

\IEEEpubid{\makebox[\columnwidth]{978-1-7281-0858-2/19/\$31.00~\copyright2019 IEEE \hfill} \hspace{\columnsep}\makebox[\columnwidth]{ }}

\maketitle
\IEEEpubidadjcol
\begin{abstract}
Massive amounts of satellite data have been gathered over time, holding the potential to unveil a spatiotemporal chronicle of the surface of Earth.
These data allow scientists to investigate various important issues, such as land use changes, on a global scale.
However, not all land-use phenomena are equally visible on satellite imagery.
In particular, the creation of an inventory of the planet's road infrastructure remains a challenge, despite being crucial to analyze urbanization patterns and their impact.
Towards this end, this work advances data-driven approaches for the automatic identification of roads based on open satellite data.
Given the typical resolutions of these historical satellite data, we observe that there is inherent variation in the visibility of different road types.
Based on this observation, we propose two deep learning frameworks that extend state-of-the-art deep learning methods by formalizing road detection as an ordinal classification task. In contrast to related schemes, one of the two models also resorts to satellite time series data that are potentially affected by missing data and cloud occlusion. Taking these time series data into account eliminates the need to manually curate datasets of high-quality image tiles, substantially simplifying the application of such models on a global scale.
We evaluate our approaches on a dataset that is based on Sentinel~2 satellite imagery and OpenStreetMap vector data.
Our results indicate that the proposed models can successfully identify large and medium-sized roads.
We also discuss opportunities and challenges related to the detection of roads and other infrastructure on a global scale.
\end{abstract}


\section{Introduction}

Scientists have been harnessing the massive openly available satellite data to study a variety of phenomena on the surface of Earth~\cite{Zhu:2019:LandsatConversation}.
Satellite data have a number of advantages compared to alternative sources.
Firstly, the relative ease of gathering these data---compared to on-site monitoring---enables the observation of land use patterns at large scales, \eg, for tracking of deforestation or agricultural monitoring. Secondly, satellite imagery are especially well-suited for tasks that require observation over time.
Unlike aerial imagery, which requires actively launching a drone or flying a plane, satellites orbit the entire Earth continuously, collecting usable data for any location on the planet within intervals of often just a few days.
As such, satellite data can be harnessed to reliably provide global snapshots of our planet over time.

\begin{figure}[t]
    \subfigure[Data]{
    \mycolorbox{\resizebox{0.42\columnwidth}{!}{\includegraphics{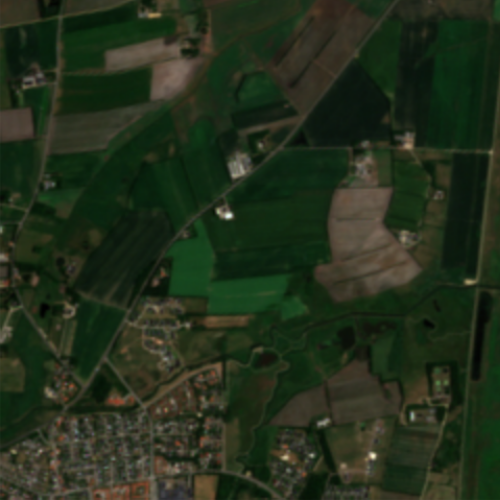}}}
    }
    \hfill
    \subfigure[Labels]{
    \mycolorbox{\resizebox{0.42\columnwidth}{!}{\includegraphics{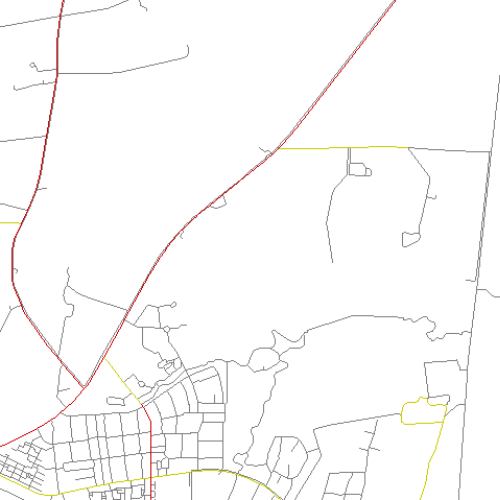}}}
    }
     \vspace{-1ex}
    \caption{Detecting roads in low-resolution satellite data: While bigger roads (red) are clearly visible, smaller roads (gray) are often at the edge of visibility. For each scene, multi-spectral images composed of several grayscale images are given; three of these images correspond to the standard RGB channels (which are shown).}
         \vspace{-2ex}
    \label{fig:example_roads}
\end{figure}

Recently, the increase of these publicly available satellite data has fueled the emergence of a wide variety of remote sensing applications.
In particular, satellite images allow the identification of details in the landscape, and recent breakthroughs in machine learning dramatically advanced the extraction of high-level information from these images~\cite{PirottiSP2016}.
However, identifying objects that are at the edge of visibility in the imagery remains a challenging task.
A particularly important class of such objects to be inventoried is the one of roads, given their impact in applications including crisis management, urban planning, or forest and land management.
For concreteness, consider that freely available imagery from Sentinel-2 satellites provides global coverage every five days at a \SI{10}{\meter} per pixel resolution.
At this spatial resolution, how distinguishable roads are varies depending on their width, which can reach sub-pixel size, see Figure~\ref{fig:example_roads}.
As a consequence, it is often the case that (historical) maps of roads for scientific analyses are still generated manually through an error-prone, time-consuming, and expensive process~\cite{laurance2018wanted}.

While current automated methods can operate effectively with sub-meter resolution data from drones, airplanes, or commercial satellites (\eg, Worldview-3), these data are very costly and unavailable over different periods and large areas.
Snapshots obtained through such methods may not allow for reconstruction of areas partially covered by clouds, for example.
Furthermore, even though the spatial resolution of publicly available satellite data is expected to improve, current methods would still be ineffective in exploiting the vast repositories of historical data for scientific analyses.
To harness these open satellite data of adequate coverage and sufficient historical sampling frequency, a fully automatic method for creating a global road inventory at any point time is highly desirable.

\newcommand{\iwidthexample}{0.14\textwidth}
\begin{figure*}[t]
    \subfigure[2016/3]{
    \resizebox{\iwidthexample}{!}{\includegraphics{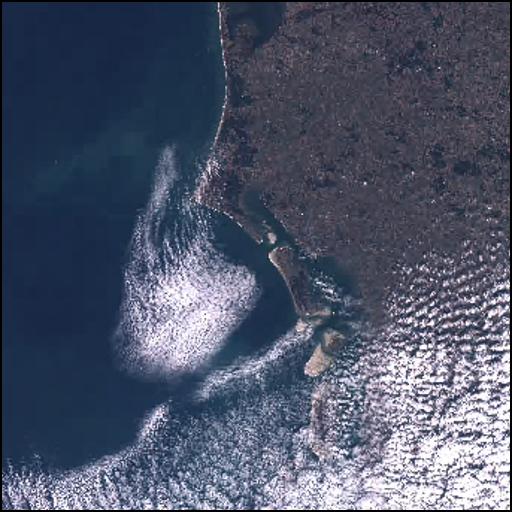}}
    }
    \hfill
    \subfigure[2016/6]{
    \resizebox{\iwidthexample}{!}{\includegraphics{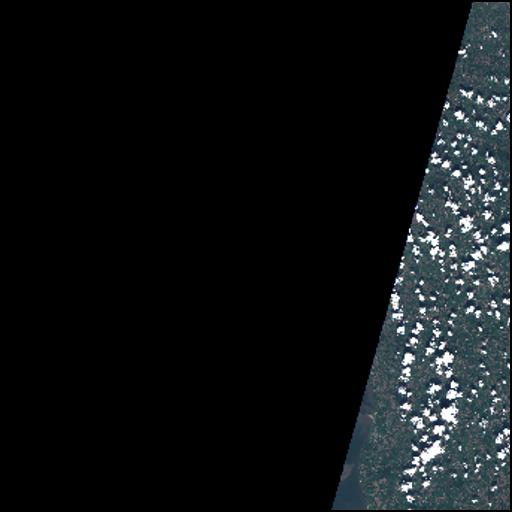}}
    }
    \hfill
    \subfigure[2016/9]{
    \resizebox{\iwidthexample}{!}{\includegraphics{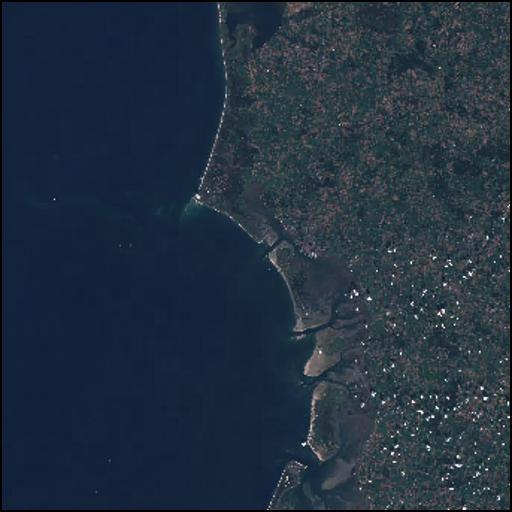}}
    }
    \hfill
    \subfigure[2016/12]{
    \resizebox{\iwidthexample}{!}{\includegraphics{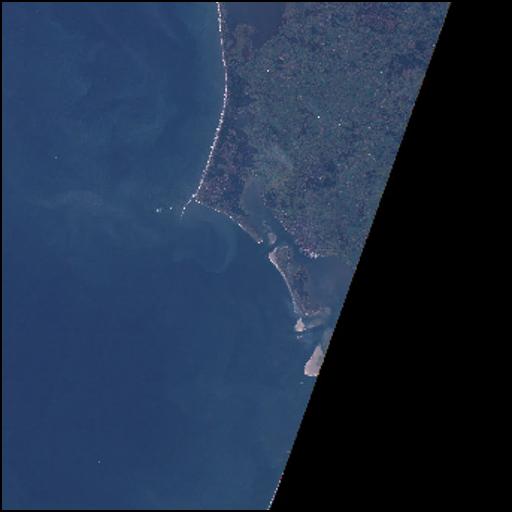}}
    }
    \hfill
    \subfigure[2017/3]{
    \resizebox{\iwidthexample}{!}{\includegraphics{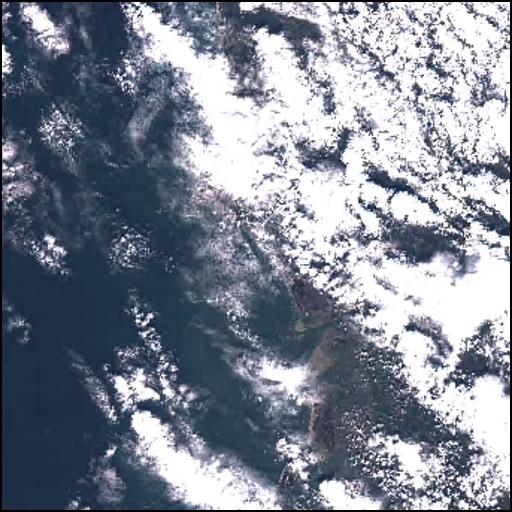}}
    }
    \hfill
    \subfigure[2017/6]{
    \resizebox{\iwidthexample}{!}{\includegraphics{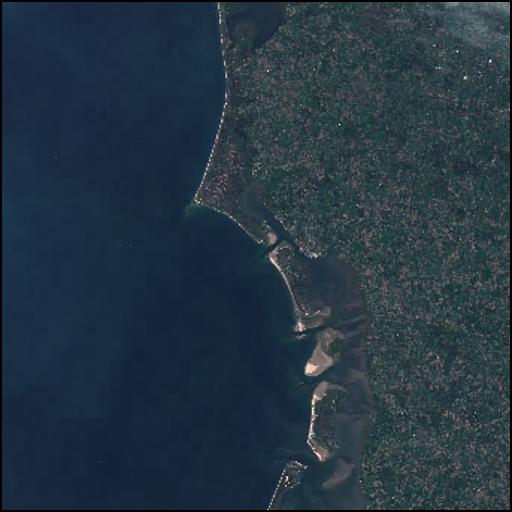}}
    }
    \hfill
    \subfigure[2017/9]{
    \resizebox{\iwidthexample}{!}{\includegraphics{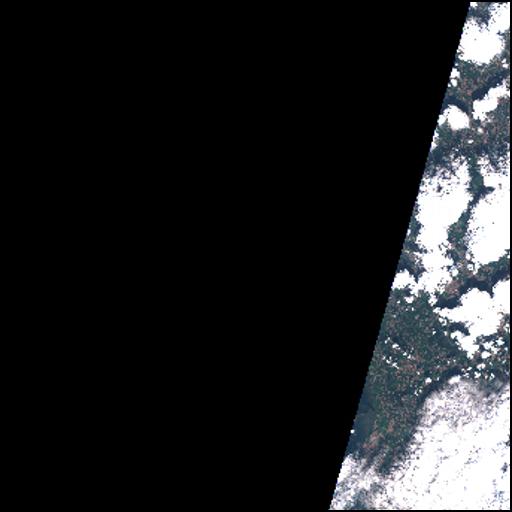}}
    }
    \hfill
    \subfigure[2017/12]{
    \resizebox{\iwidthexample}{!}{\includegraphics{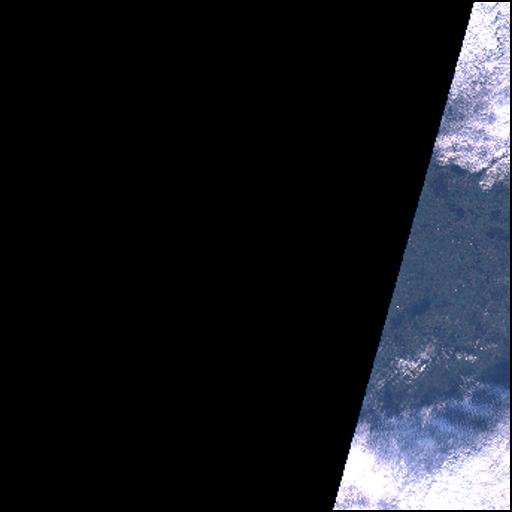}}
    }
    \hfill
    \subfigure[2018/3]{
    \resizebox{\iwidthexample}{!}{\includegraphics{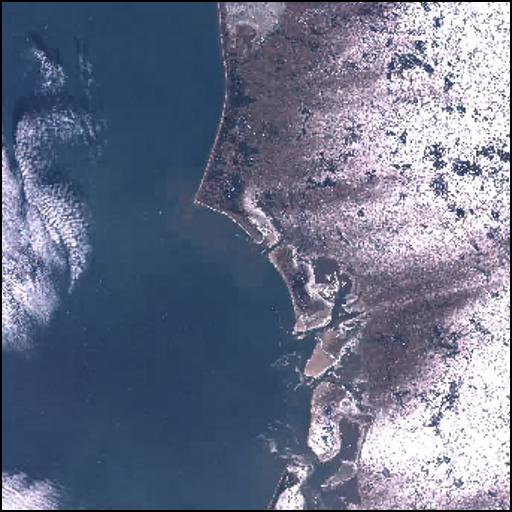}}
    }
    \hfill
    \subfigure[2018/6]{
    \resizebox{\iwidthexample}{!}{\includegraphics{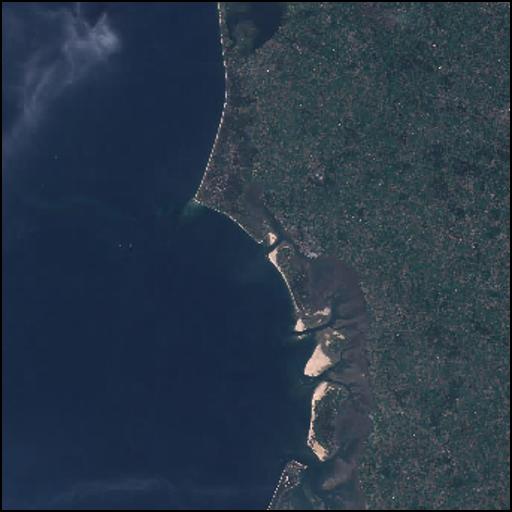}}
    }
    \hfill
    \subfigure[2018/9]{
    \resizebox{\iwidthexample}{!}{\includegraphics{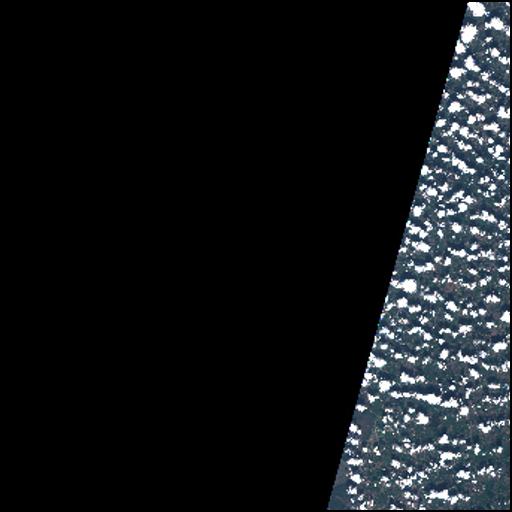}}
    }
    \hfill
    \subfigure[2018/12]{
    \resizebox{\iwidthexample}{!}{\includegraphics{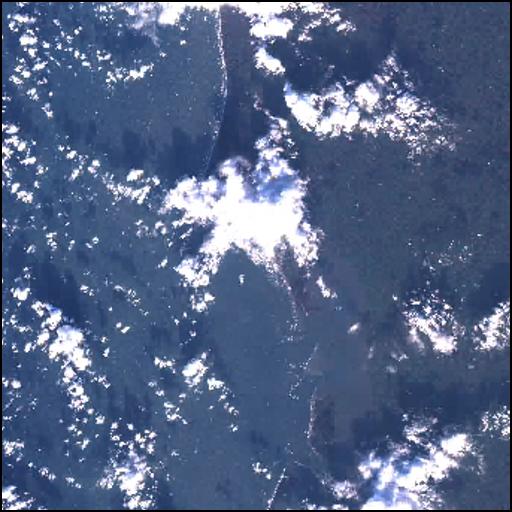}}
    }
    \vspace{-1ex}
    \caption{
    Sentinel~2 time series image data (RGB) given for a certain tile (T32UMG)~\cite{sentinel-2}.
    It can be seen that the image quality is both affected by clouds as well as by missing data.
    The models considered in this work either operate on images of ``high quality'' (i.e., very few clouds) or on such sequences of potentially ``low quality'' images to automatically extract road information.
    In the latter case, since the relevant information are automatically extracted from the individual images, even regions with a high cloud coverage can be processed.
    This is an important ingredient for extracting road maps on a global scale from the available massive amounts of data.}
    \vspace{-2ex}
    \label{fig:example_tile}
\end{figure*}

This work addresses the task of extracting roads from open satellite data repositories by extending state-of-the-art segmentation architectures commonly used in this context.
In particular, we propose two deep road extraction architectures by building upon two observations.
Firstly, we observe that more precise label information can be used to capture barely visible roads, \ie, label data with a higher spatial resolution than the satellite input images to detect finer road structures.
The first model is developed based on this observation and relies on curated images, \eg, images exhibiting a low cloud coverage, where scenes are arguably of high quality.
Secondly, we observe that a time series of images of potentially lower quality (\eg, affected by clouds or missing data) is available for all scenes, see Figure~\ref{fig:example_tile}.
These time series can be exploited by the second model---which also takes more precise label information into account---to automatically extract the relevant road information from multiple different images that are relatively close in time.
The second model significantly simplifies the use of open satellite data towards global-scale scientific analyses, since only low quality assumptions (not all images are completely cloud covered) have to be fulfilled for the data that are processed.
The models are capable of reliably extracting large and medium-sized roads (\eg, motorways, primary, and secondary roads) as well as small roads (\eg, roads in residential areas) with consistently good performance.
Given these results, we believe that the proposed models can also be promising candidates for the detection of other hardly visible objects on the surface of our planet.

\section{Background}\label{sec:background}
This section reviews background related to the satellite data as well as to state-of-the-art approaches for road extraction.

\subsection{Earth Observation \& Geospatial Data}\label{subsec:earth-observation&geospatial-data}
Up until the early 2000s, imagery of high spatial resolution was mostly obtainable through commercial acquisition of data products from satellites or by use of drones and airplanes.
As mentioned previously, the availability of global-scale coverage from these data sources is limited, and the associated data products can be cumbersome to process (\eg, due to cloud cover).
Moreover, the temporal resolution of these sources can leave much to be desired, posing an obstacle for longitudinal scientific analyses of changes in land use.

This situation changed when imagery started to be provided freely to the public~\cite{Zhu:2019:LandsatConversation} with satellite data exhibiting a spatial resolution of \SI{15}{\meter} to \SI{30}{\meter} available on a global scale~\cite{WULDER20122}.
The Sentinel-2 mission has since revolutionized the monitoring of Earth~\cite{rs9090902}.
The mission provided images at a high frequency (new image every five days), which ensures that a dense time series for each surface location is available.
Further, the mission's data products went up to \SI{10}{\meter} spatial resolution per pixel, enhancing visibility.
Since this data is free and already preprocessed, the usage was ensured.
As such, Sentinel-2 achieved spatial and temporal coverage that is, so far, unique.

Despite the immense advances of the Sentinel-2 mission, challenges remain in the development of models to process its data products.
A number of the images are occluded by clouds, which is usually dealt with by disregarding the affected images or by applying masks on the affected areas.
Unfortunately, these approaches drastically reduce the number of images available to an object detection model and necessitate some degree of manual intervention for use of the data.

Previous work on pixel-based classification approaches indicates that there is potential to leverage Sentinel-2 images to classify sub-pixel landscape features, such as roads~\cite{rs8060488} or ships~\cite{Heiselberg2016}.
However, while traditional machine learning approaches, such as random forests~\cite{Breiman2001} or support vector machines~\cite{Vapnik1995}, have been routinely applied to detect land cover classes~\cite{PirottiSP2016}, surprisingly little work has been conducted on detecting these hardly visible objects in Sentinel-2 data via modern deep learning techniques.

A fundamental challenge in pursuing a detection approach based on deep learning for objects of varying visibility, such as roads, is the availability of adequate label data.
A potential source of these label data, investigated in the present work, is volunteered geographical information (VGI)~\cite{flanagin2008credibility}.
With VGI, the crowd employs a platform to contribute to the curation of a common geospatial dataset.
The quality of such a dataset then depends on the review processes and reputation of the platform as well as on the commitment, redundancy in local expertise, and number of individual contributors.
A prime example of a high-quality VGI platform is OpenStreetMap (OSM)~\cite{haklay2008openstreetmap}, which has been in operation since the early 2000s.
Due to its large community, OSM includes data of global coverage.
Moreover, OSM provides versioning capabilities, thus enabling the selection of label data that can match particular snapshots in time.
The latter is important to align the label data with the satellite data at large scales.
Notably, we assume that changes in roads are slow enough within short intervals, \eg, one week to one month, that an appropriate snapshot selection can be made without difficulty.

The representation of VGI in OSM follows a long tradition of data modeling in GIS~\cite{goodchild1992geographical}.
In particular, geographical features are encoded as vector data, providing a resolution-resilient simplification of geographical reality.
Roads are represented as sequences of line segments associated with attribute data.
Importantly, these data include the type of road, \eg, motorway, trunk, residential, among many others.
While these data are employed by traditional GIS applications for matching or mapping geospatial data, they can be a useful source for categorization of roads into degrees of visibility and for thus providing more precise labels.

\subsection{Road Detection}
While OSM data are often available and relatively complete for densely populated areas, it is often missing precise and up-to-date label information for sparsely populated areas.
This makes automatic infrastructure detection schemes essential for a variety of applications including urban planning or crisis management.
The detection of roads or other infrastructure elements is typically addressed in two steps.
In the first step, a segmentation into, \eg, ``road'' and ``no road'' pixels of the input data is generated.
In the second step, the segmented images are usually converted to vector data~\cite{ijgi7030084,Mattyus_2017_ICCV}.
The focus of this work is on the first step, \ie, on segmenting satellite imagery into the corresponding classes.

One of the first approaches for the automatic detection of roads from satellite data was proposed by Bajcsy and Tavakoli~\cite{BajcsyRuzena1976CRoR}.
In recent years, deep learning methods~\cite{LeCunBH2015} have been used to identify roads, mostly based on high-resolution image data gathered by drones or airplanes.
For instance, Minh and Hinton~\cite{MnihVolodymyr2010LtDR} use a deep learning approach to extract road information from aerial imagery with a high resolution.
This approach has been adapted by Zhang and Liu~\cite{ZhangZhengxin2018REbD}, who propose a network architecture called Deep ResUNet that yielded slightly better results.
Recent deep learning approaches rely on high-resolution data (\eg, with a resolution of \SI{0.5}{\meter} per pixel) and OSM label information to generate models for the automatic extraction of roads or buildings~\cite{bonafilia2019building,ijgi7030084,Mattyus_2017_ICCV}.
To account for cloud cover, Ru{\ss}wurm and K\"orner~\cite{russwurm2018multi} processed sequences of satellite data for land cover classification with a recurrent convolutional approach.

Radoux~\etal~\cite{rs8060488} have analyzed the potential of detecting very small objects given Sentinel data and pointed out that ``the Sentinel-2 roads detection limit was of \SI{3}{\meter}''.
While deep learning has been used in the context of imagery with a sub-meter resolution, surprisingly little work has been conducted on extracting road information from such satellite images with a relatively low resolution (which are available through the Landsat and Sentinel missions).
One approach aiming at the extraction of roads based on Sentinel data has been proposed by Abdelfattah and Chokmani~\cite{8127809}, who resort to a feature extraction scheme to detect rural and desert roads.
However, no state-of-the-art deep learning architecture has been considered for the extraction of roads from low-resolution satellite time series data.

\section{Deep Road Extraction}
\label{sec:models}
This section will introduce two road detection models that are adapted to the specific needs of the data at hand;
their empirical performance will be analyzed in Section~\ref{sec:experiments}.
To fit the models, we are given a set \( T = \{(\x_1, \y_1), \ldots, (\x_\npattern,\y_\npattern) \} \subset \X \times \Y\) consisting of images~\(\x_i \in \mathbb{R}^{\iwidth\times \iheight\times\ndim} \) with associated labels \(\y_i \in \mathbb{R}^{\iwidth\times \iheight \times \nclasses}\) for the individual pixels, where $\iwidth$ and $\iheight$ specify the width and height of an image and were $\nclasses$ specifies the number of classes.
Each image instance is composed of multiple bands given for a single time steps (\eg, $\ndim=13$ bands) or of multiple bands (\eg, 13) given for several time steps (\eg, 12, leading to $\ndim=13\cdot12$ channels), see the appendix for details regarding the particular data considered.
The goal of the overall learning process is to find suitable models of the form \(\model : \X \rightarrow \Y\) that output pixel-wise classifications for new, unseen input data.

\subsection{Model Architectures}\label{subsec:model-architectures}
\begin{figure}
    \centering
    \resizebox{1.0\columnwidth}{!}{\includegraphics{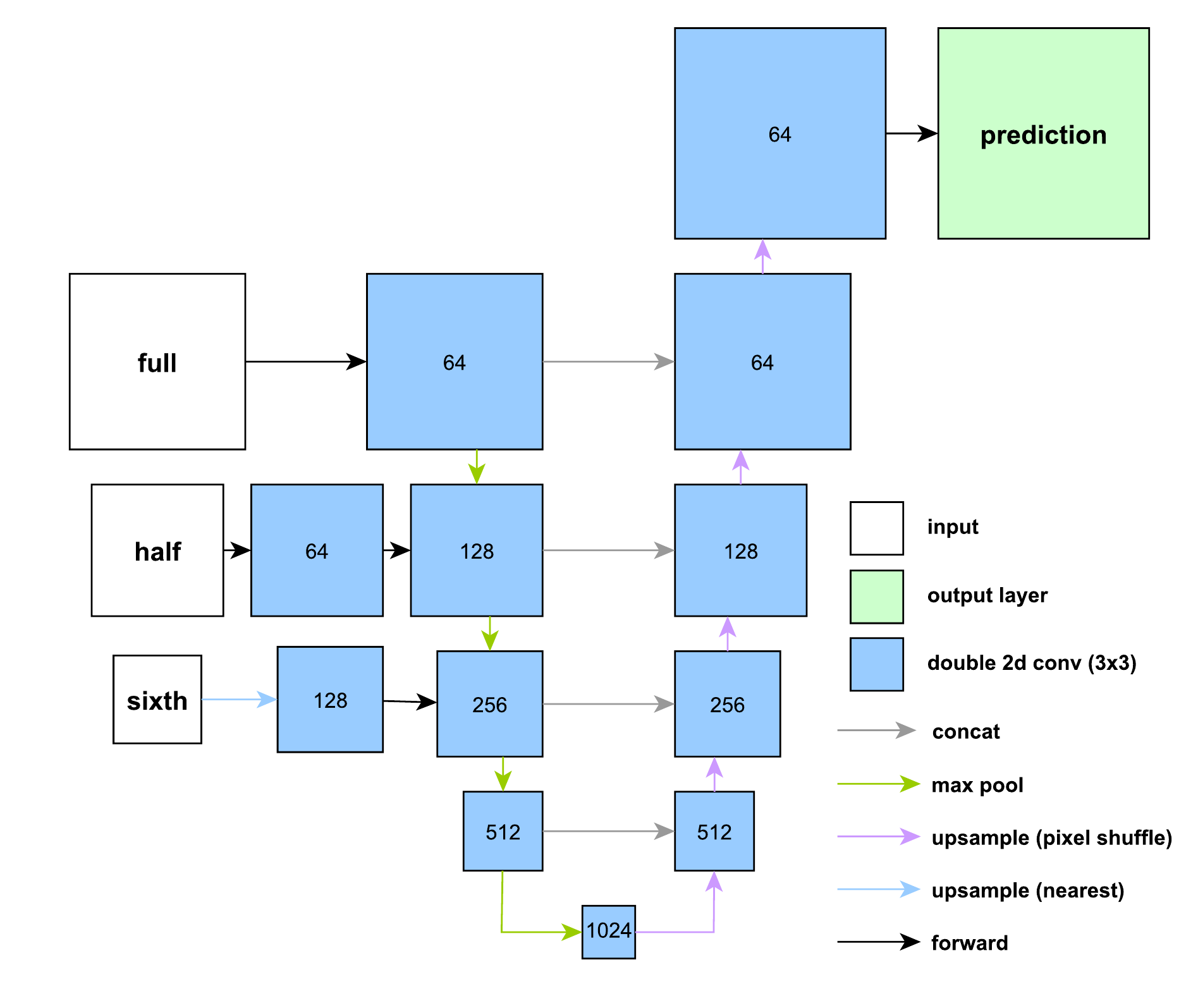}}
    \vspace{-1ex}
    \caption{
    \unetplus: Segmentation model that operates on single ``high-quality'' input scenes.
    In the case of Sentinel-2 data, the input data is given in terms of different bands with a \SI{10}{\meter} to \SI{60}{\meter} resolution.
    The labels can be provided with a higher spatial resolution (e.g., \SI{5}{\meter} or better).
    This more precise label information is taken into account by considering an additional upsample and double convolution block after the decoding part.
    The numbers within boxes indicate number of feature maps at the output of a block.
    }
    \vspace{-2ex}
    \label{fig:model1}
\end{figure}
Both models depict extensions of the popular \emph{U-Net}~architecture proposed by Ronneberger~\etal~\cite{RonnebergerFB15}.
The architecture exhibits an encoding and a decoding part.
The encoding part typically consists of a sequence of two consecutive convolution layers with \(3\times3\) filter followed by a pooling layer, resulting in a set of feature maps (sometimes called \emph{bottleneck features}).
At each level of the encoder, the spatial dimensions of the features are reduced to half their sizes, while the number of filters for the convolution layers is doubled, thus increasing the receptive field of the convolution operator and thereby allowing the network to extract increasingly context-rich information.
The decoding part is structured as if the encoder is reversed, whereby the pooling layers are replaced by upsampling layers.
The decoders final feature maps have the same shape as the input data.
To obtain the final segmentation results, \texttt{softmax} or \texttt{sigmoid} activation functions are typically used~\cite{RonnebergerFB15}.
The different levels of the decoder part are also connected to the corresponding levels of the encoder part via so-called skip connections, which transfer feature maps from the encoding to the decoding part by concatenating the feature map outputs of the encoder and with the feature maps of the decoder part~\cite{RonnebergerFB15,LongSD14}.
After each convolution layer, except for the final output layer, batch normalization and the \texttt{ReLU} activation function are applied.

\subsubsection{Single High-Quality Scene}
The first network architecture (called \unetplus) extends the U-Net architecture by exhibiting a higher spatial resolution for the labels compared to the input images, see Figure~\ref{fig:model1}.
More precisely, we make use of the fact that the vector data is available with a higher spatial resolution (\eg, with a \SI{5}{\meter} resolution or even higher compared to the \SI{10}{\meter} resolution of Sentinel~2 data).
The model incorporates the more precise labeled data via a final upsampling layer that outputs data with a higher resolution than the input data (and matching the one of the provided labels).
This simple yet crucial modification allows the network to learn label road segmentations exhibiting a higher resolution than the input images, which is important in case fine details are to be extracted (\eg, small roads).
For the upsampling layers, we employ---instead of using deconvolution---the pixel-shuffle approach that was introduced to create super-resolution images, which decreases checkerboard artifacts~\cite{shi2016real}.
As we will show in the experimental evaluation, the induced model yields more accurate predictions compared to its direct competitor that resorts to label information with a normal resolution.

As input data, the model receives a single satellite image of high quality (\ie, few to no clouds), which is composed of different bands.
Since the bands are available in different resolutions, we incorporate the different bands at their corresponding positions into the architecture (there are three resolutions in our scenario; ``full'', ``half'', and ``sixth'', corresponding to images with a resolution of \SI{10}{\meter}, \SI{20}{\meter}, and \SI{60}{\meter}, respectively).
The input labels are provided as rasterized vector data, see the appendix for the details.


\subsubsection{Sequences of Low-Quality Scenes}
\begin{figure}[t]
    \subfigure[Data]{
    \mycolorbox{\resizebox{0.42\columnwidth}{!}{\includegraphics{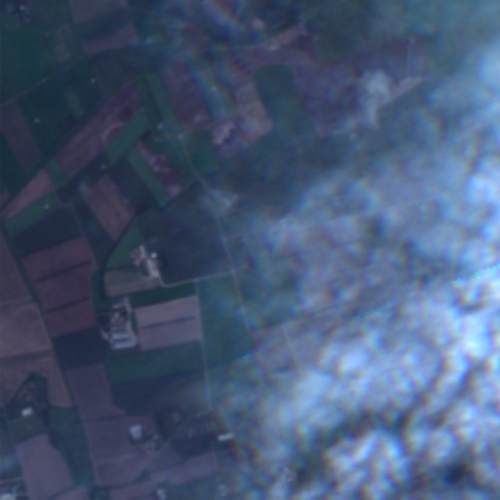}}}
    }
    \hfill
    \subfigure[Predictions]{
    \mycolorbox{\resizebox{0.42\columnwidth}{!}{\includegraphics{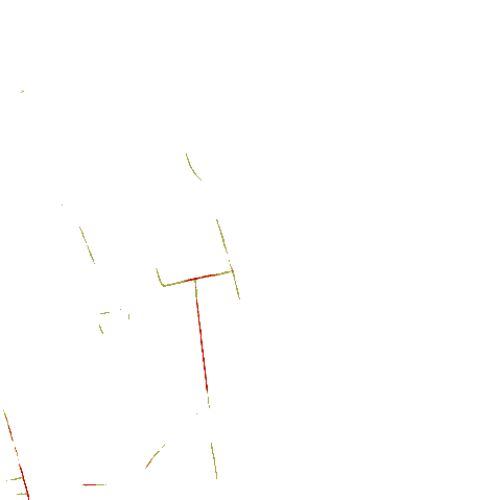}}}
    }
    \vspace{-1ex}
    \caption{
    In case parts of the input data are occluded by clouds (even only slightly), a model that processes single scenes might not be able to detect roads anymore.
    In practice, this is usually handled by processing ``similar'' scenes (e.g., an image that was taken in the same month).
    However, this usually complicates the automatic processing of large amounts of data.
	Also, in the worst case, no completely cloud-free patch might be available for a particular scene.
	}
    \vspace{-2ex}
    \label{fig:example_clouds}
\end{figure}
The first model can successfully detect roads in case single high-quality images are available (\ie, images that are not affected by clouds).
However, such images might not be available for the target scene;
typically, parts of a patch are affected by clouds, which can significantly reduce the prediction performance of the model, see Figure~\ref{fig:example_clouds} for an illustration.
Simply using ``surrogates'' (\eg, training data from snow-covered January, but test data in green July) might lead to a worse performance due the dataset shifts induced by seasonal changes.

\begin{figure}
    \centering
    \resizebox{1\columnwidth}{!}{\includegraphics{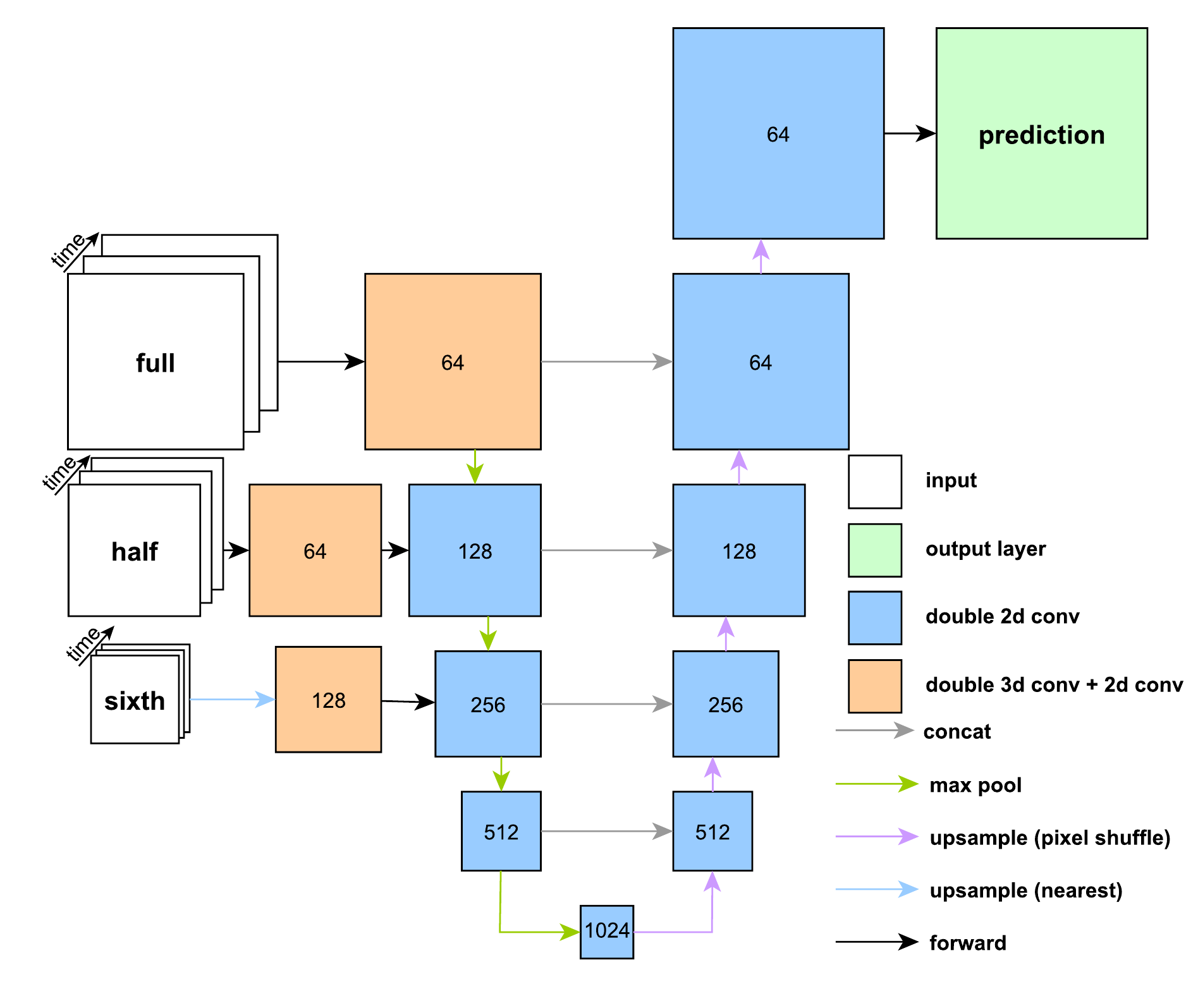}}
    \vspace{-1ex}
    \caption{
    \unetplustimevolumetric: Segmentation model that operates on a sequence of input scenes, each composed of multiple bands.
    The network automatically identifies the cloud-free parts of the different input bands and fuses this information for the subsequent layers to predict roads.
    Note that this not only handles scenarios that are affected by clouds, but also cases where roads are occluded by, e.g.,  trees (in the summer).
    }
    \vspace{-2ex}
    \label{fig:model2}
\end{figure}

Our second architecture (called \unetplustimevolumetric) operates on time series data of potentially ``low quality'', where each image can possibly be affected by clouds or missing data, see again Figure~\ref{fig:example_tile}.
The network architecture, shown in Figure~\ref{fig:model2}, receives, for each time step, a set of bands (as before).
The bands and time steps are combined via two consecutive volumetric (3d) convolution layers~\cite{3dconv} with a \(3\times 3 \times 3\) filter and are then flattened and aggregated via a \(1\times 1\) 2d convolution layer to a single time step.
Note that the 3d convolution layers are only employed after the input layer and then aggregated to save computational resources. We also divided the number of neurons for the 3d convolution layers by 4 to have comparable model sizes (\num{43527300} parameters for \unetplustimevolumetric\ compared to \num{43045764} parameters for \unetplus).
This allows the network to learn temporal dependencies present in the data, while significantly reducing the computational requirements compared to a network with 3d convolution layers throughout the whole architecture.
As before, more precise label information is taken into account via one or more final upsampling layers.
As we will show in the experimental evaluation, this model can successfully deal with sequences of images of very low quality.
Compared to the \unetplus\ model, this model significantly simplifies the processing of large areas since basically no manual selection of individual input tiles has to be conducted.

\subsection{Training and Inference}
Typically, road detection and related tasks are handled as standard classification tasks (\eg, based on the two classes ``road'' and ``no road'' predicted for each pixel).
For the data at hand, this is, however, problematic since quite a large amount of small roads are hardly visible or even not visible at all in the satellite data with the given resolution (\eg, tiny roads with a width of less than \SI{3}{\meter}).
This complicates the training because the network will try to find roads at locations where it is practically impossible to detect them in the data.

Instead, we treat the road detection task as an ordinal classification problem with the following four classes (\(c=4\)): \noroad, \smallroad~(potentially not visible), \mediumroad~(most likely visible), and \bigroad~(should be visible), see again Figure~\ref{fig:example_roads} and the appendix for details regarding the labels.
Modeling the road detection task this way avoids that false predictions get penalized too heavily, \ie, the different classification levels render the loss function to be more suited for the given problem.
Formally, we learn \(c-1\) classifiers (with \(c\) being the number of classes) such that each classifier also ``contains'' the previous classifiers' target, learning to the final layers of the networks having \(c-1\) outputs.
This is similar to a one-hot encoding, but in order to predict class \(c_i\), all previous outputs have to be above the prediction threshold as well (\eg, above \num{0.5}).
During training, each classifier is trained as a binary classifier based on a user-defined loss function (see below).
At test time, to reach a final decision, the sigmoid activation function is applied, whose output is discretized through thresholding (\num{0} or \num{1}), leading to \(c\) outputs.
Finally, the model predictions are cascaded and evaluated.
For instance, given $c=4$, the output vector $(1,0,0)$ would be mapped to the class \smallroad, $(1,1,0)$ to the class \mediumroad, and $(1,1,1)$ to the class \bigroad.
Note that $(1,0,1)$ would be mapped to \smallroad.
No roads would be predicted for outputs whose first entry is zero (\eg, (0,0,0), but also (0,1,0)).
Thus, to predict a class, all previous classes need to ``activated'' as well.
This encoding was 
introduced by Frank and Hall~\cite{ordclas} and has recently been applied in the context of medical segmentation problems~\cite{ordseg}.


The segmentation task at hand is highly unbalanced with class \noroad\ dominating the other three classes.
For this reason, we resort to the Tversky loss function defined as
\[\tverskyloss(p, \hat{p}) = \frac{p\cdot\hat{p}}{ p\cdot\hat{p} + \beta(1-p) + (1- \beta ) p \cdot( 1 - \hat{p} )  }, \]
with label~\( p \in \{ 0, 1 \} \) and prediction~\( \hat{p} \in [0,1]\) (for all experiments, $\beta$ was set to \( \beta = 0.7 \)).
As optimizer, we resort to AdamW~\cite{loshchilov2018decoupled} with learning rate $lr=3e-4$, L2-regularization with $l2=0.0005$, and cosine annealing with warm restarts (standard parameters \(T_0=1\) \(T_{mult}=2\))~\cite{loshchilov-ICLR17SGDR}.
We also conduct several data augmentation steps during training (horizontal flops, vertical flips, instance min-max normalization).

The overall performance was assessed based on the \emph{precision} $\frac{tp}{tp+fp}$, the \emph{recall} $\frac{tp}{tp+fn}$, and the induced \emph{F1-score}
\begin{equation*}
    2 \times \frac{ precision \times recall} {precision + recall},
\end{equation*}
where $tp$ is the number of true positives, $fp$ the number of false positives, and $fn$ the number of false negatives.
These scores are computed for each of the individual classes, leading to, \eg, one F1-score for each of the three classes \smallroad, \mediumroad, and \bigroad.
In the evaluation we also consider the \emph{Jaccard-Index} (intersection-over-union):
\begin{equation*}
    \frac{ precision \times recall} {precision + recall},
\end{equation*}
which is penalizing single bad predictions more than the \emph{F1-score}.
So, if model~A has a better \emph{Jaccard-Index} then model~B, its worst predictions are better than the ones of model~B.

All models presented in this work were trained for 500 epochs and the best-performing models w.r.t. the average F1 score over the road classes were selected based on a validation set (10\% of the training data).

\section{Experiments}
\label{sec:experiments}
All models were implemented in Python~(3.6) using PyTorch~(1.0) and were trained on a training set consisting of labeled patches;
their final performance was measured on a separate test set. A detailed description of the dataset can be found in the Appendix. 
For the road labels, one is given the four classes mentioned above.
Note that the vector layer is converted to a rasterized image by treating a line segment as the ``middle'' of the road.
Consequently, the segmentation task is to label the middle of the roads, allowing for identification and classification of roads at a finer level. 
We compare the performances of the following models:
\begin{itemize}[leftmargin=*,topsep=0pt,partopsep=0pt]
    \item \unet: The baseline model, which resorts to a standard U-Net architecture that is commonly used in the context of road detection scenarios~\cite{ijgi7030084}.
    The model operates on data from a single timestamp of high quality (very few clouds in the test set; the tenth scene in the sequence of the twelve scenes, corresponding to June 2018).
    The resolution of the labels is \(10\times 10\) meters, matching the highest one of the satellite images given.
    \item \unetplus: A direct extension of the \unet\ baseline model that receives label information with a higher resolution (\(5\times 5\) meters), as presented in Section~\ref{sec:models}.
    \item \unetplustimeflat: An extension of the \unetplus\ model, which processes the images given for all the twelve timestamps in a ``flat'' fashion, \ie, all the input images are treated as independent feature maps. Notably,
the images given for a particular band (\eg, B01, see appendix) are treated as ``independent'' feature maps.
    \item \unetplustimevolumetric: The second model presented in Section~\ref{sec:models}, which resorts to volumetric convolutions to merge the imagery per band given for the twelve timestamps.
\end{itemize}
Note that the '\texttt{+}' indicates that labels with a higher spatial resolution are incorporated.
We ensured that all models have roughly the same amount of learnable parameters.
The first two models operate on data from a single timestamp only, whereas the last two models operate on data from the sequence of twelve timestamps.

In the remainder of this section, we provide results showing that the modifications had an impact on the model quality as well as an overall model comparison, followed by a brief discussion of opportunities and challenges in this context.

\subsection{More Precise Labels}\label{subsec:more-precise-labels}
\begin{figure}[t]

    \subfigure[Data]{
    \mycolorbox{\resizebox{0.42\columnwidth}{!}{\includegraphics{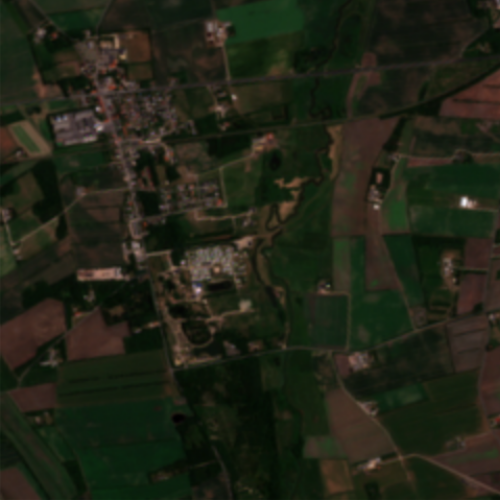}}}
    }
    \hfill
    \subfigure[Labels]{
    \mycolorbox{\resizebox{0.42\columnwidth}{!}{\includegraphics{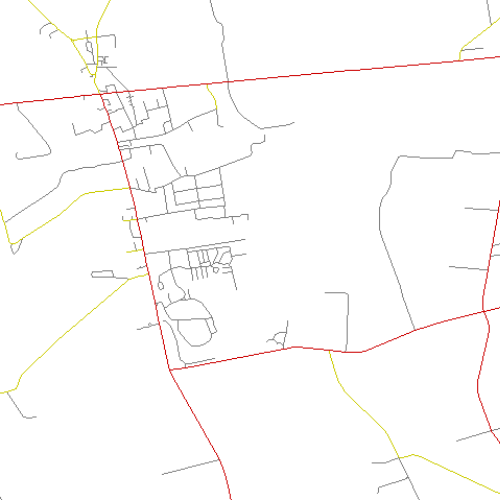}}}
    }
    \subfigure[\unet]{
    \mycolorbox{\resizebox{0.42\columnwidth}{!}{\includegraphics{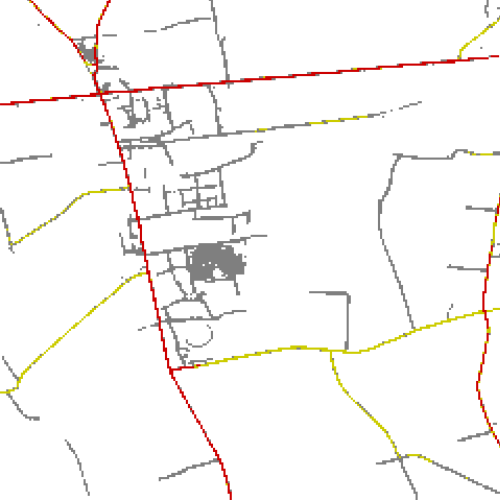}}}
    }
    \hfill
    \subfigure[\unetplus]{
    \mycolorbox{\resizebox{0.42\columnwidth}{!}{\includegraphics{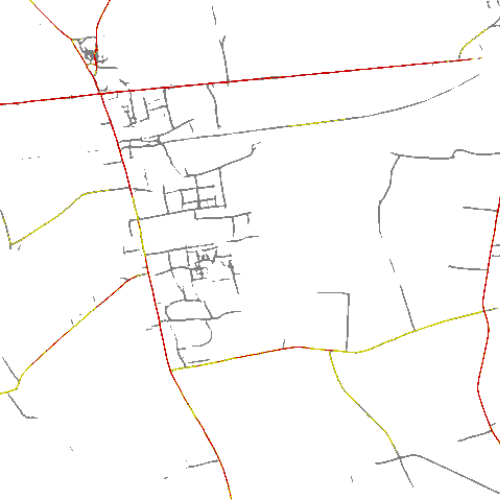}}}
    }

  \vspace{-1ex}
    \caption{
    Comparison of the baseline model \unet\ and our \unetplus\ extension.
    It is shown that the latter model can detect roads at a finer level compared to the baseline.
    }
    \vspace{-2ex}
    \label{fig:resolution}
\end{figure}

A small yet crucial modification incorporated in both model architectures presented in Section~\ref{sec:models} is the use of road labels with a higher spatial resolution than the input images.
This information is incorporated via an additional upsampling and convolution layer after the decoding part of the network.
The general idea behind this modification is that the labels typically used in this context are too coarse to identify fine details, which are still visible in the data.
In Figure~\ref{fig:resolution}, a comparison between the baseline model (\unet) and its direct extension \unetplus\ is given.
It can be seen that the \unetplus\ model can extract the roads with a higher precision than the \unet\ model, but also finds roads where no labels exist.
Note that the road labels are given at a sub-pixel resolution compared to the input imagery.
This clearly indicates the potential of models processing label information that are given with a higher spatial resolution than the one of the input imagery.

\subsection{Time Series Data}\label{subsec:time-series-data}
Both the \unet\ and the \unetplus\ model work on imagery of high quality given for a single timestamp (in this case, a tile collected in June 2018).
While such models are generally able to yield satisfying results given data that are not affected by clouds, they usually fail in the case when data of low quality are given during the inference phase, see again Figure~\ref{fig:example_clouds}.
Typically, one addresses this problem by retrieving ``surrogate'' patches that are not affected by clouds (\eg, by considering a tile collected in May 2018).
However, such a surrogate might not be available.
Also, resorting to surrogates might introduce shifts due to the vegetation varying in the course of a year.
Finally, identifying patches that are potentially affected by clouds generally requires manual preprocessing steps (\eg, the application of a cloud detection model) and ignoring data that is only slightly affected by thin clouds can be suboptimal.

\begin{figure}[t]

    \subfigure[Data]{
    \mycolorbox{\resizebox{0.42\columnwidth}{!}{\includegraphics{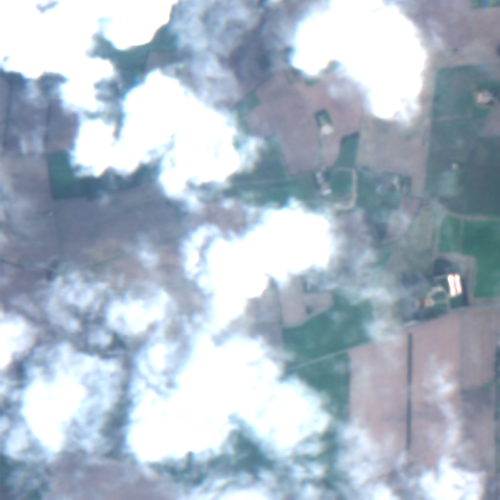}}}
    }
    \hfill
    \subfigure[Labels]{
    \mycolorbox{\resizebox{0.42\columnwidth}{!}{\includegraphics{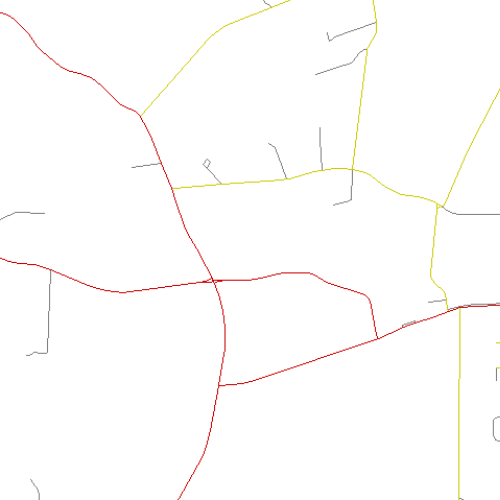}}}
    }
    \subfigure[\unetplus]{
    \mycolorbox{\resizebox{0.42\columnwidth}{!}{\includegraphics{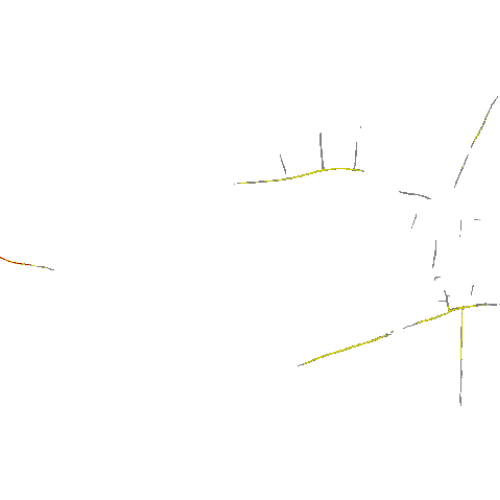}}}
    }
    \hfill
    \subfigure[\unetplustimevolumetric]{
    \mycolorbox{\resizebox{0.42\columnwidth}{!}{\includegraphics{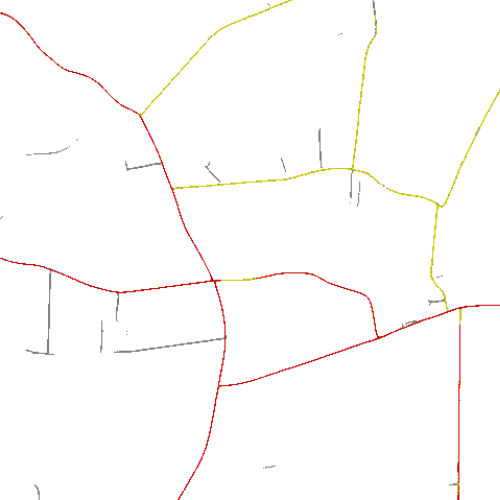}}}
    }
    \vspace{-1ex}
    \caption{
    Comparison of \unetplus\ and \unetplustimevolumetric\ models.
    It can be seen that the second model that operates on sequences of images can successfully detect the road infrastructure.
    The \unetplus\ model, which only operates on data given for a single timestamp, cannot detect any roads in the affected parts of the image.
    }
    \vspace{-2ex}
    \label{fig:cloud_comparison}
\end{figure}

\begin{figure}
    \centering
    \resizebox{1\columnwidth}{!}{\includegraphics{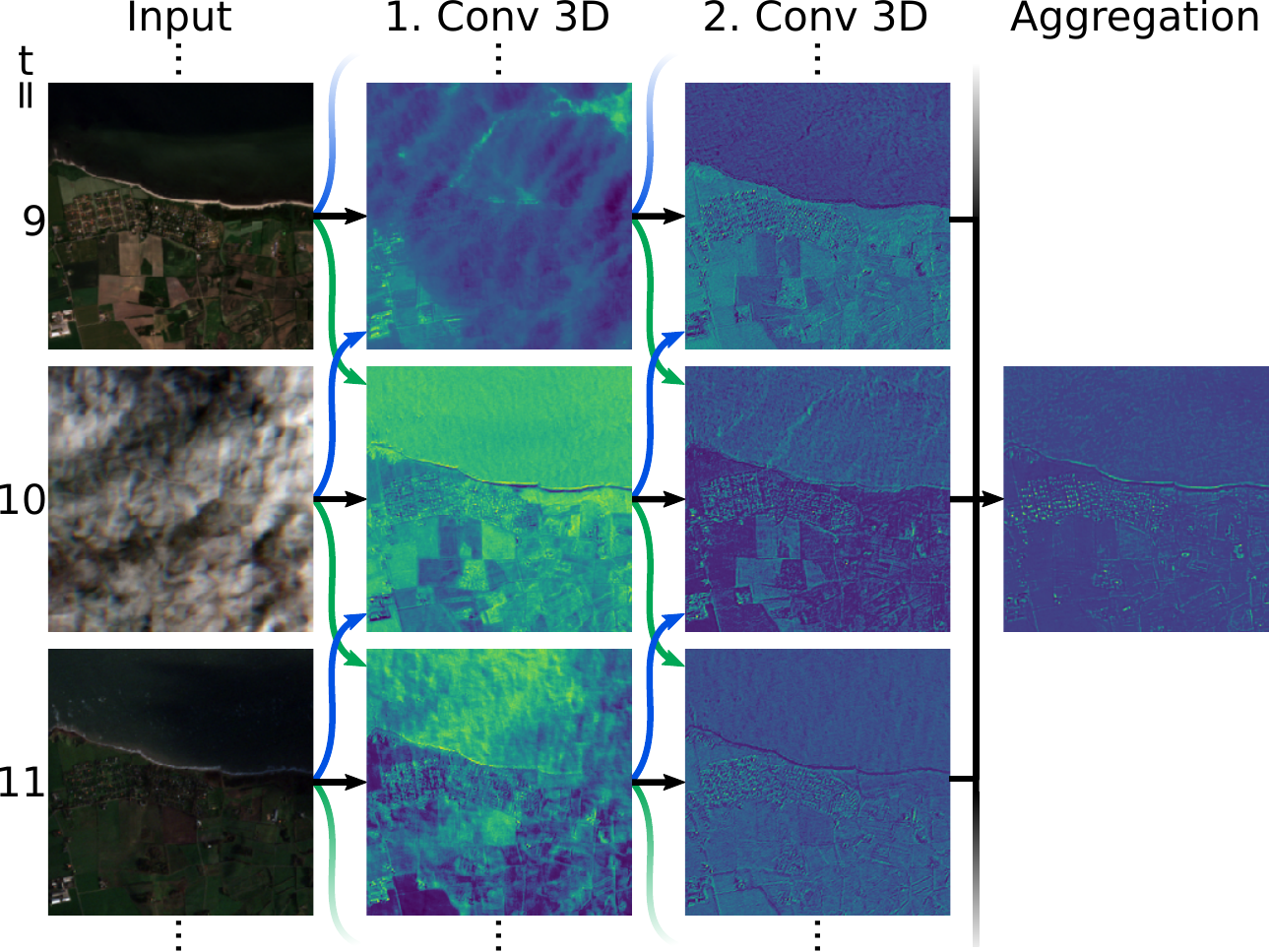}}
    \caption{
    The first feature maps of a \unetplustimevolumetric\ model instance.
    Notice that after the second volumetric convolution (3x3x3 filter), the clouds are no longer visible in the feature maps.
    These ``cleaned'' feature maps are combined to a single feature map via the aggregation layer (1x1 2d convolution layer).
    }
    \vspace{-2ex}
    \label{fig:conv}
\end{figure}

These issues are addressed by the two model implementations that operate on the whole sequence of imagery.
In Figure~\ref{fig:cloud_comparison}, a comparison between the \unetplus\ and the \unetplustimevolumetric\ model is given for a test patch that is affected by clouds.
As expected, it can be seen that the \unetplus\ model cannot infer the road infrastructure, while the \unetplustimevolumetric\ can successfully detect most of the roads.
In Figure~\ref{fig:conv}, the time feature fusing process conducted by the \unetplustimevolumetric\ model is illustrated.

\subsection{Model Comparison}\label{subsec:model-comparison}
\begin{table*}
    \sisetup{
    table-align-uncertainty=true,
    separate-uncertainty=true,
    }
    \renewrobustcmd{\bfseries}{\fontseries{b}\selectfont}
    \renewrobustcmd{\boldmath}{}
    \caption{
    Comparison of the different models considered in this work.
    The best results are shown in bold. \label{table:results}
    }
    \vspace{-2ex}
    \center

    \resizebox{1.0\textwidth}{!}{
    \begin{tabular}[t]{lll
    S[table-format=1.3,detect-weight,mode=text]
    @{\hspace{.1cm}}
    S[table-format=1.3,detect-weight,mode=text]
    @{\hspace{.1cm}}
    S[table-format=1.3,detect-weight,mode=text]
    @{\hspace{.1cm}}
    S[table-format=1.3,detect-weight,mode=text]
    @{\hspace{.1cm}}
    S[table-format=1.3,detect-weight,mode=text]
    @{\hspace{.1cm}}
    S[table-format=1.3,detect-weight,mode=text]
    @{\hspace{.1cm}}
    S[table-format=1.3,detect-weight,mode=text]
    @{\hspace{.1cm}}
    S[table-format=1.3,detect-weight,mode=text]
    @{\hspace{.1cm}}
    S[table-format=1.3,detect-weight,mode=text]
    @{\hspace{.1cm}}
    S[table-format=1.3,detect-weight,mode=text]
    @{\hspace{.1cm}}
    S[table-format=1.3,detect-weight,mode=text]
    @{\hspace{.1cm}}
    S[table-format=1.3,detect-weight,mode=text]
    @{\hspace{.1cm}}
    }

        \toprule

        model & \(5\times 5\)m & time &  \multicolumn{3}{c}{Jaccard-Index} & \multicolumn{3}{c}{Precision} & \multicolumn{3}{c}{Recall} & \multicolumn{3}{c}{F1-score} \\
        \cmidrule(lr){4-6} \cmidrule(lr){7-9} \cmidrule(lr){10-12} \cmidrule(lr){13-15}
        & & & \multicolumn{1}{l}{small} & \multicolumn{1}{l}{mid.} & \multicolumn{1}{l}{big} & \multicolumn{1}{l}{small} & \multicolumn{1}{l}{mid.} & \multicolumn{1}{l}{big} & \multicolumn{1}{l}{small} & \multicolumn{1}{l}{mid.} & \multicolumn{1}{l}{big} & \multicolumn{1}{l}{small} & \multicolumn{1}{l}{mid.} & \multicolumn{1}{l}{big} \\
        \midrule
        \unet & no& single & 0.164 & 0.182 & 0.281 & 0.196 & 0.211 & 0.327 & \bfseries 0.499 & \bfseries 0.564 & 0.668 & 0.281 & 0.307 & 0.439 \\
        \unetplus & yes & single & 0.208 & 0.253 & \bfseries 0.414 & 0.294 & 0.349 & 0.519 & 0.418 & 0.481 & \bfseries 0.671 & 0.345 & 0.404 & \bfseries 0.585 \\
        \unetplusshift & yes & shift & 0.023 & 0.008 & 0.023 & 0.142 & 0.063 & 0.238 & 0.027 & 0.009 & 0.025 & 0.045 & 0.015 & 0.045 \\
        \unetplustimeflat & yes & flat & 0.208 & 0.250 & 0.375 & 0.297 & 0.330 & \bfseries 0.553 & 0.413 & 0.509 & 0.537 & 0.345 & 0.400 & 0.545 \\
        \unetplustimevolumetric & yes & 3d & \bfseries 0.217 & \bfseries 0.264 & 0.392 & \bfseries 0.317 & \bfseries 0.351 & 0.550 & 0.408 & 0.515 & 0.576 & \bfseries 0.357 & \bfseries 0.417 & 0.563 \\

        \bottomrule
    \end{tabular}
    }
\end{table*}
An overall model comparison is presented in Table~\ref{table:results}.
In addition to the aforementioned models, we also considered a variant of the \unetplus\ model, named \unetplusshift, that was trained on patches taken in June 2018 but applied to patches in September 2018 (most of the patches were affected by clouds).
It can be seen that the \unet\ baseline exhibits inferior Jaccard-Indices and lower F1-scores compared to \unetplus, \unetplustimeflat, and \unetplustimevolumetric, whereas it generally exhibits higher recalls.
This is as expected since the \unet\ model operates on data with a higher spatial resolution of \(10\times 10\) meters, leading to too many pixels being labeled as roads by the model, see again Figure~\ref{fig:resolution}.\footnote{For comparison purposes, the predictions of the \unet\ model are upsampled (nearest neighbor method) to match the spatial resolution of its competitors.}
Especially when fine details are supposed to be detected in the imagery, this is an undesired behavior as the precision is very low.
Another obvious observation is that the \unetplusshift\ model yields very bad overall performance, since 
most of the patches to which it is applied are affected by clouds.

\begin{figure*}
    \centering
    \includegraphics[width=1.0\textwidth]{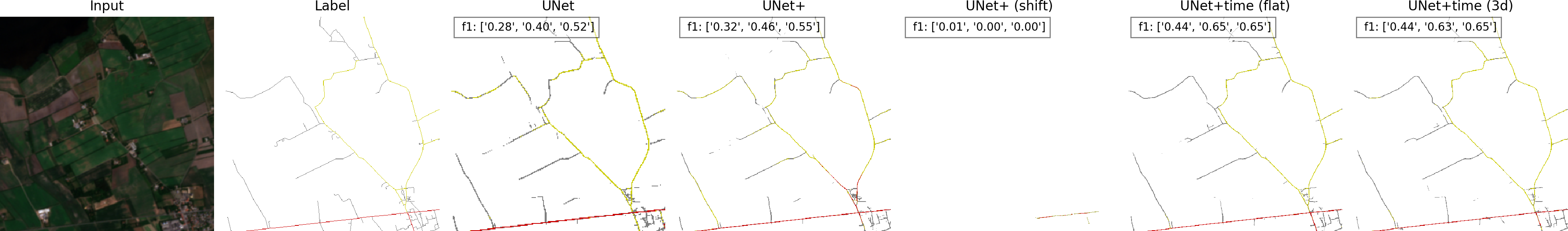}
    \includegraphics[width=1.0\textwidth]{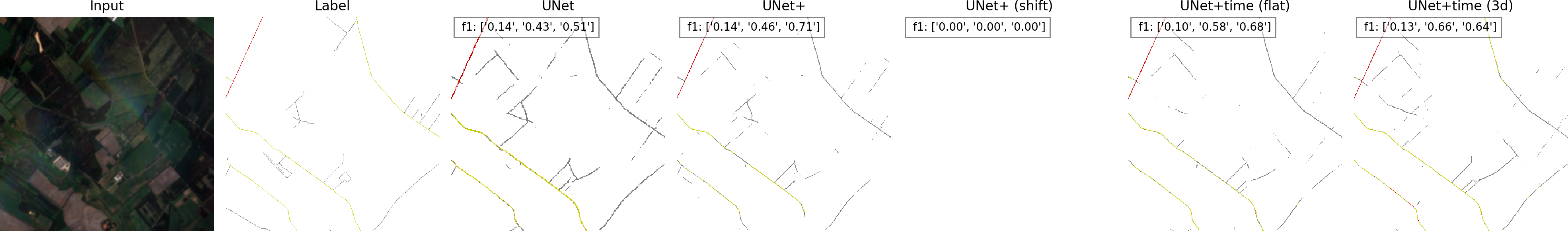}
    \includegraphics[width=1.0\textwidth]{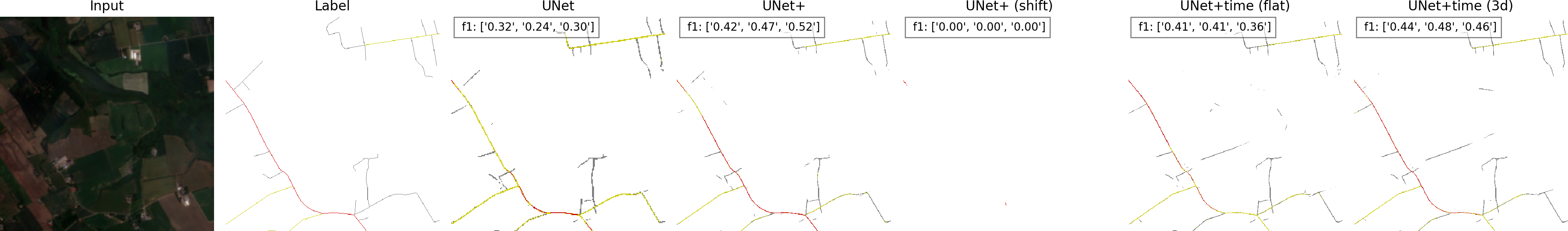}
    \includegraphics[width=1.0\textwidth]{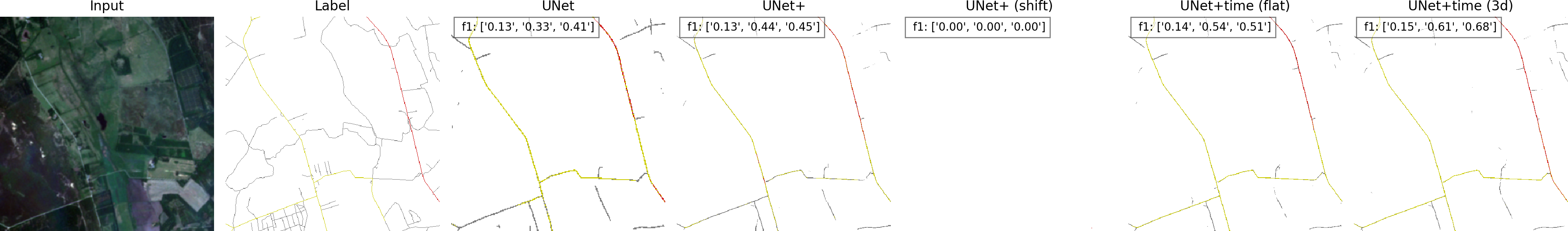}
    \includegraphics[width=1.0\textwidth]{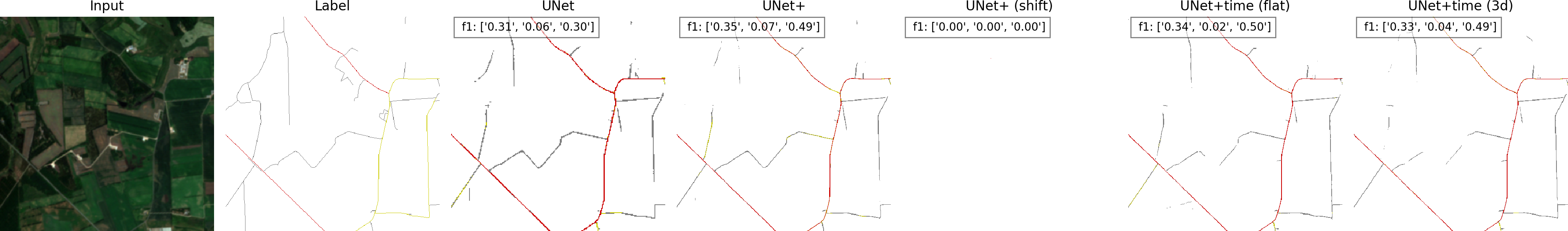}
    \caption{
    Results for a random subset of test patches.
    It can be seen that the models with a high spatial resolution can successfully generate the labels at a high resolution (as expected).
    }
    \label{fig:examples_predictions}
    \vspace{-2ex}
\end{figure*}

Both the \unetplus\ and the \unetplustimevolumetric\ model show a very similar performance.
As illustrated above, the latter model can, however, automatically retrieve the relevant information from a sequence of images that are potentially affected by clouds or missing data.
This is an essential ingredient when processing large amounts of satellite data, since the overall process is significantly simplified.
Finally, it can be seen that the \unetplustimevolumetric\ model exhibits a better performance than the \unetplustimeflat\ model instance, indicating that the volumetric convolutions help to structure the feature maps at the beginning.
In Figure~\ref{fig:examples_predictions}, the classification results for a random subset of patches are provided.
Especially the two models \unetplus\ and \unetplustimevolumetric\ yield satisfying results, particularly for large and medium-sized roads.
In some cases, all models fail to identify the small roads.
However, such roads are almost invisible in the data at hand.

\subsection{Opportunities \& Challenges}\label{subsec:opportunities&challenges}
\begin{figure*}[t]
    \centering
    \mycolorbox{\resizebox{.96\textwidth}{!}{\includegraphics{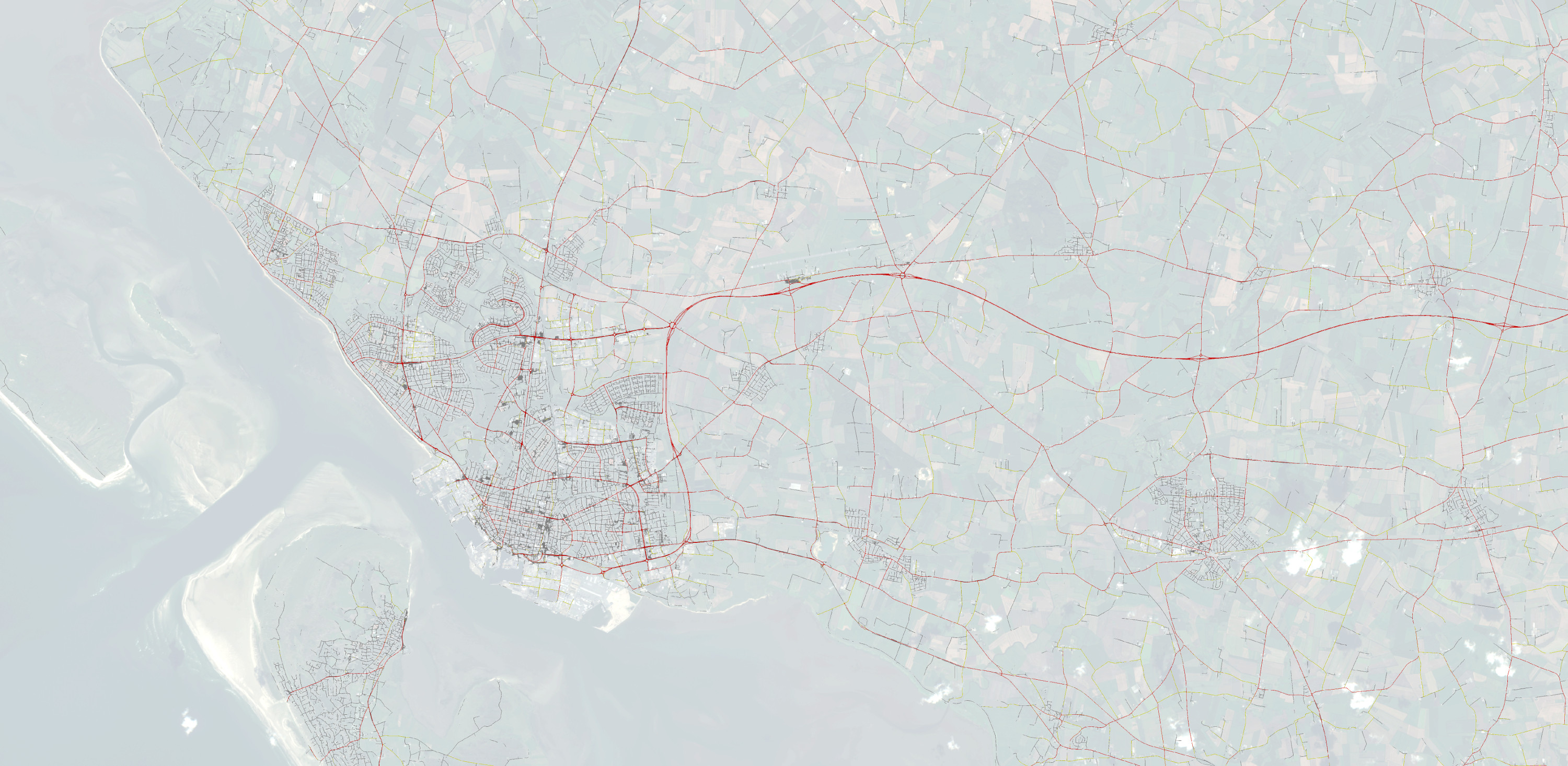}}}
    \caption{
    Map generated based on predictions produced for individual patches (an area in the west of Denmark is shown).
    }
    \vspace{-2ex}
    \label{fig:map}
\end{figure*}

The results shown above display that the models proposed in this work can successfully detect large and medium-sized roads given low-resolution satellite data with a spatial resolution of \(10\times 10\) meters.
Naturally, small roads are hardly visible in such data.
The models presented can identify those with consistently good performance. 
Furthermore, the \unetplustimevolumetric\ model significantly simplifies the processing of large amounts of satellite data since almost no manual preprocessing steps have to be conducted (\eg, generation of cloud-free mosaics).
While the model operates on patches with a relatively small size, large maps can easily be generated via a sliding window approach:
To create consistent maps, one can partition the larger scene into patches with a large overlap (\eg, with a stride of half the patch size).
This ensures that the predictions made for the individual patches can be successfully merged without any artifacts at the boundaries (the output of the model is deterministic as long as the receptive field for a pixel does not exceed the patch).
An illustration of such a map is shown in Figure~\ref{fig:map}, which shows an area in the west of Denmark.
This map can serve as the input for follow-up approaches that, for instance, again vectorize the results.

Various challenges can still be addressed in the future.
The results reported above clearly indicate that fine details and roads can successfully be extracted.
One of the main reasons for prediction errors is naturally given by the limited resolution of the available satellite data, \ie, given images with a spatial resolution of \(10\times 10\) meters, extracting roads with a width of five meters or even less might be impossible.
Another limitation, which might be dealt with in future, relates to the label information:
The labels used for generating the models were not optimal yet.
Firstly, the road labels might be one or a few pixels off, which complicates both the training as well as the evaluation.
Secondly, the road attributes considered to group the roads into the three classes were not always in line with the reality, \ie, some major roads were labeled as \smallroad, whereas some small roads were labeled as \bigroad.
One way to improve this would be to incorporate precise, error-free labels indicating the particular type and the width of the road.
Alternatively, the models considered can be further adapted to deal with such label noise.
Finally, aiming at analyses on a global scale, one needs to have training data covering all parts considered in the inference phase
or otherwise, transfer learning approaches have to be employed to lessen the negative side-effects caused by domain shifts.

\section{Conclusions}\label{sec:conclusions}
We have addressed the task of extracting (hardly visible) roads from low-resolution satellite imagery.
Our deep learning approaches resort to label data that is given at a higher resolution than the input images, which allows the networks to detect fine details in satellite images with consistently good performance. 
We also consider time series of satellite images (potentially partially affected by clouds or missing data) to extract the desired road infrastructure.
This significantly simplifies the application of such models on global scales, since only low quality assumptions have to be fulfilled for the data that are processed. 
We also sketched challenges related to the detection of small roads and outlined potential directions for future research. 


\subsection*{Acknowledgements}
We would like to thank the Microsoft AI for Earth program for providing cloud computing resources that have facilitated parts of this work.

We acknowledge support from the Danish Industry Foundation through the Industrial Data Analysis Service (IDAS).
\appendix\label{sec:appendix}

The training and test sets are extracted from a dataset that is based on Sentinel~2 images~\cite{rs9090902}, which are organized via so-called tiles.
For each such tile, one is given image data that are continuously collected by the satellite with a revisit time of five days, see Figure~\ref{fig:example_tile} for an illustration.
For each tile and timestamp, the data are given in terms of several (grayscale) images, which correspond to so-called bands and which exhibit a spatial resolution of $10\times10$ (B02, B03, B04, B08), $20\times20$ (B05, B06, B07, B8A, B11, B12), and $60\times60$ (B01, B09, B10) meters per pixel, respectively.
For the experiments conducted in this work, image data of six tiles were considered (32UMG, 32UNG, 32UPG, 32VMH, 32VNH, 32VNJ).
For each tile, twelve timestamps were considered in the period from 2016 till 2018 (per year: March, June, September, and December), yielding $12\cdot13$ grayscale images per tile (Level-2A products were used).
The corresponding road labels were based on vector data gathered by the
\emph{OpenStreetMap}~(OSM)~\cite{OpenStreetMap} project.\footnote{Downloaded via \url{https://download.geofabrik.de} as shape files.}
For each road segment, one is given attribute data indicating the type of road.
This data was used to partition the roads into the following three classes:
\begin{itemize}
    \item \texttt{big}: All road segments with one of the following OSM labels: 'motorway', 'motorway\_link', 'primary', 'primary\_link', 'secondary', 'secondary\_link', 'tertiary', 'tertiary\_link'
    \item \texttt{medium}: All road segmented with OSM label 'unclassified'.
    \item \texttt{small}: All remaining road segments.
\end{itemize}
The vector data was rasterized with a spatial resolution of $10\times10$ (for the baseline U-Net model) and $5\times5$ (for all other models) meters, respectively.
For each road, the middle/center of that road is characterized by the road segment in the vector data (\ie, no information related to the particular width is given).
For each such tile, \num{1000} random patches were extracted.
The patch sizes for the different input bands varies from $40 \times 40$ (for the bands with a resolution of \SI{60}{\meter}) to $240 \times 240$ (for the bands with a resolution of \SI{10}{\meter}).
The label information was given as $240 \times 240$ patches (baseline U-Net model) and $480  \times 480$ (other models).
The induced patches were stored in binary (NumPy) files, exhibiting a total data volume of about 50~GB\@.
The patches of the 32VHM tile were used for the test set;
all remaining patches were used for training the models.

\bibliographystyle{IEEEtran}
\bibliography{biblio}

\end{document}